\documentclass[11pt,a4paper]{article}
\usepackage[hyperref]{naaclhlt2018}
\usepackage{times}
\usepackage{latexsym}
\usepackage{graphicx}
\usepackage{subfig}
\usepackage{enumitem}
\usepackage{colortbl}
\usepackage{amsmath}
\usepackage{amsfonts}
\usepackage[export]{adjustbox}
\usepackage{etoolbox}
\AtBeginEnvironment{quote}{\singlespacing\small}
\setlist[itemize]{topsep=0.5pt}
\usepackage{url}

\aclfinalcopy 

\title{Using Sentiment Induction to Understand Variation \\ in Gendered Online Communities}

\author{Li Lucy \\
  Symbolic Systems Program \\
  Department of Computer Science \\
  Stanford University \\
  {\tt lucy3@stanford.edu} \\\And
  Julia Mendelsohn \\
  Department of Linguistics \\
  Department of Computer Science \\
  Stanford University \\
  {\tt jmendels@stanford.edu} \\}
\date{}

\begin{document}
\maketitle
\begin{abstract}
  We analyze gendered communities defined in three different ways: text, users, and sentiment. Differences across these representations reveal facets of communities' distinctive identities, such as social group, topic, and attitudes. Two communities may have high text similarity but not user similarity or vice versa, and word usage also does not vary according to a clearcut, binary perspective of gender. Community-specific sentiment lexicons demonstrate that sentiment can be a useful indicator of words' social meaning and community values, especially in the context of discussion content and user demographics. Our results show that social platforms such as Reddit are active settings for different constructions of gender. 
\end{abstract}

\section{Introduction}

Social groups can be described by many factors, such as the demographics of its participants or its physical location. To detect sociolinguistically significant groups, linguists have built upon the concept of \textit{communities of practice}, which are characterized by their participants' shared actions, beliefs, values, and language styles \cite{eckert1992think, eckert2006communities}. This concept initially emerged to understand the complex interplay between language and gender, and has been applied to study social identities in numerous communities \cite[e.g.][]{eckert1989jocks,mendoza1996muy,hall2009boys}. While previous variationist work has primarily studied traditional physical communities, we focus instead on online ones.

Online communities have been shown to form collective linguistic norms, which give rise to a rich amount of language variation across communities, even on the same website \citep{danescu2013no,yang2015putting}. One of the largest content aggregator and discussion platforms, Reddit, contains thousands of unique communities, known as \textit{subreddits}. These subreddits vary in topic, such as r/sport and r/history, content type, such as r/pics and r/videos, and format, such as the Q\&A style of r/IamA and the narratives on r/confession. Online communities such as subreddits are often characterized by language use and user membership \citep{hamilton2016inducing,datta2017identifying,bamman2014gender,martin2017community2vec}. 

Sociolinguists have primarily analyzed phonological and syntactic variables \cite{eckert2012three}, though some have studied lexical variables \cite[e.g.][]{wong2005reappropriation}. Previous computational work also focuses on lexical variation \cite{bamman2014gender}. We approach variation from a new direction, where we examine the salient semantic dimension of sentiment to understand \textit{how} users use the same words to convey different meanings. We also create representations for subreddits that encode text and user membership to situate insights gained from sentiment-based representations and to understand the intersection of speaker identity (user), content (text), and affect (sentiment). This paper focuses on explicitly gendered subreddits, which cater towards masculine- or feminine-identifying groups. 

We provide two main contributions:
\begin{itemize}[leftmargin=.15in]
\setlength\itemsep{0em}
  \item[1)] Salient aspects of social group identities, such as gender, can produce low user overlap in communities sharing similar topics. 
  \item[2)] Sentiment-based representations of communities can reveal a type of variation across social groups that word-choice alone cannot. An in-depth study of words' sentiments in gendered subreddits reveals patterns of how linguistic resources construct a wide array of gendered identities in the online sphere.
\end{itemize}

\section{Previous Work}

\begin{table*}
\centering
\footnotesize
\begin{tabular}
{|c|>{\raggedright\arraybackslash}m{11.5cm}|}
\hline
Subreddit & Description \\
\hline
r/actuallesbians & ``A place for cis and trans lesbians, bisexual girls, chicks who like chicks..."\\
\hline
r/askgaybros & ``Where you can ask the manly men for their opinions on various topics."\\
\hline
r/mensrights & ``For those who wish to discuss men's rights and the ways said rights are infringed upon."\\
\hline
r/askmen & ``A semi-serious place to ask men casual questions about life, career, and more."\\
\hline
r/askwomen & ``Dedicated to asking women questions about their thoughts, lives, and experiences."\\
\hline
r/xxfitness & ``For women and gender non-binary redditors who are fit, want to be fit..."\\
\hline
r/femalefashionadvice & ``A subreddit dedicated to learning about and discussing women's fashion."\\
\hline
r/malefashionadvice & ``Making clothing less intimidating and helping you develop your own style."\\
\hline
r/trollxchromosomes & ``A subreddit for rage comics and other memes with a girly slant."\\
\hline
\end{tabular}
\caption{Gendered communities with descriptions from their sidebars or subreddit search listings.}
\label{tbl:gender_subred}
\end{table*}

The study of online communities is highly interdisciplinary, spanning machine learning, natural language processing, social network analysis, communications, and sociolinguistics. Twitter, online news, and other websites have been a rich source of data for computational social scientists, and Reddit in particular has been of interest to much previous work \cite{althoff2014ask, kumar2018community, newell2016user, jaech2015talking, hamilton2017loyalty}. Research in this area expands beyond quantitative measures, by comparing results with social science theories and employing qualitative thinking to highlight trends of individual words and communities \cite{bamman2014gender, danescu2013no, zhang2017community, althoff2014ask}. 

In particular, the characterization and identification of gender is a common use case for online data. Language is often used to infer demographics of users, especially in classification tasks that tend to find clear distinctions between men and women \cite[e.g.][]{burger2011discriminating, argamon2007mining, schler2006effects, rao2010classifying}. Previous work have found strong patterns of gender differences in language \cite{newman2008gender,mulac2001empirical}. For example, \citet{volkova2013exploring} showed that the sentiment of words, hashtags, and emoticons vary between men and women on Twitter and used gender-dependent features to improve sentiment classification of tweets. Our work aims to look beyond a straightforward divide between men and women, particularly because gender is not a fixed biological variable, but rather a dynamic, social one that is actively created and reinforced through repeated behaviors \cite{butler1988performative,nguyen2014gender, herring2006gender}. 

Language is used to simultaneously co-construct multiple identities, so the plethora of gendered identities that emerge from communities of practice may substantially differ from mainstream stereotypes of ``femininity" and ``masculinity" \cite{eckert1989jocks,mendoza1996muy,hall2009boys}. On Twitter, \citet{bamman2014gender} investigated cross-community lexical variation and the varied ways of constructing gender identities. They clustered users based on bag-of-words representations of their posts, and the resulting clusters corresponded not only to topical interest but also gender. Some clusters had language patterns that were orthogonal to expected language differences between men and women, demonstrating diversity in gendered language styles. 

We compare text, user, and sentiment representations of communities by their predictions of similarity and identify cases where these predictions agree or disagree. \citet{pavalanathan2017multidimensional} suggested that subreddits with similar topics can have dissimilar user groups due to differences in preferred interactional styles. \citet{datta2017identifying} introduced a method for finding misalignments of inferred user-based and text-based networks on Reddit. They found that pairs with high text but low user similarity tend to be communities that conflict (such as political subreddits) as well as communities with hierarchical relationships (such as a niche subreddit with a more generic one). High user but low text similarity suggested a single overarching community scattered across multiple subreddits. We assessed~\citet{datta2017identifying}'s $z^2$-score method in section \ref{text_users}, but found that its results hid important misalignments. 

We use domain-specific lexicon induction techniques for creating sentiment-based subreddit representations. Previous work has built or adapted word embeddings or scores to flexibly encode semantic dimensions or specific domains, and some have applied their techniques to social media communities \cite{rothe2016ultradense, yang2015putting,hamilton2016inducing}. \citet{rothe2016ultradense}'s \textsc{Densifier} model involves dense word embeddings created by mapping generic word embeddings into meaningful subspaces. These learned embeddings may even contain a single dimension, which can act as labels for an induced lexicon, and are best applied to corpora with several billion tokens, which is far greater than any subreddit that we study. 

While \citet{rothe2016ultradense} learned lexicons for generic domains such as news and Twitter, \citet{hamilton2016inducing}'s \textsc{SentProp} method solves a similar task across fine-grained domains, including Reddit communities. They applied a label propagation method to 250 Reddit communities, and found a wide range of sentiment variation. For example, \textit{insane} is negative in r/twoxchromosomes but positive in r/sports, while \textit{soft} shows the opposite pattern. \textsc{SentProp} was the most suitable approach for our purposes due to its ability to operate on smaller datasets. We extend this work by examining how vector representations created from subreddit-specific sentiment lexicons compare to text-based and user-based representations, with gendered communities as a case study. 

\section{Data}

In order to gain a broad perspective of how gendered subreddits relate to each other and other communities within the larger Reddit context, we consider data from subreddits with 50,000+ subscribers, as provided in a user-curated list\footnote{Available \href{https://www.reddit.com/r/ListOfSubreddits/wiki/listofsubreddits}{here}.}. These subreddits span topics ranging from plants to cryptocurrency, and provide a glimpse into Reddit's long tail of diverse niche communities, which is a primary draw to the platform \cite{newell2016user}. The most popular subreddits tend to be part of a set of ``default" subreddits to which users have historically been auto-subscribed ~\cite{newell2016user,datta2017identifying}. To focus on more niche and non-artificially inflated communities, we filtered out about 50 default subreddits based on lists in r/defaults created during May 07 2014, May 26 2016, and March 26 2017. 

We took the top 400 remaining subreddits and used their comments created between May 2016 and April 2017. The vast majority of these subreddits contain between $10^7$ and $10^8$ tokens, with r/politics (the largest) containing over 764 million tokens to r/accidentalwesanderson (the smallest) containing over 7 million. From these subreddits we manually selected nine subreddits with clearly gender-oriented names (Table \ref{tbl:gender_subred}).

\section{Approach}

\begin{table}
\centering
\footnotesize
\begin{tabular}{|l|m{5cm}|}
\hline
Positive & love, loved, loves, awesome, nice, amazing, best, fantastic, correct, happy \\
\hline
Negative & hate, hated, hates, terrible, nasty, awful, worst, horrible, wrong, sad \\
\hline
\end{tabular}
\caption{Positive and negative Twitter seed words from \citet{hamilton2016inducing} }
\label{tbl:sentiment_seed}
\end{table}

\subsection{Text \& User Representations}

To provide a basis of comparison for our sentiment-based representations, we created term frequency-inverse document frequency (tf-idf) vectors for each subreddit using user and unigram frequencies. Here, we define subreddit user frequencies as the number of times a user comments to a specific subreddit. For unigram or user $t$ in subreddit $d$, its tf-idf weighted frequency is

$$w_{t,d} = (1 + \log \mathit{tf}_{t,d})\log(N/\mathit{df}_t),$$

where $\mathit{tf}_{t,d}$ is the frequency of $t$ in $d$, $\mathit{df}_t$ is the number of subreddits in which $t$ appears, and $N$ is the total number of subreddits. 

We filtered out rare users and bots, with $1 < \mathit{df}_t \leq 380$ for users, and filtered out rare words and stop words, with $5 < \mathit{df}_t \leq 380$ for unigrams. We used truncated singular value decomposition (SVD) to reduce these vectors to 100 dimensions and normalized them to each have a unit norm \cite{scikit-learn,halko2011finding}.

\subsection{Sentiment Representations}

We induced community-specific sentiment lexicons using the \textsc{SentProp} method introduced by \citet{hamilton2016inducing}. This framework was demonstrated to perform well on moderately sized domains of $10^7$ tokens, which matches the majority of our subreddits. 

\textsc{SentProp} begins by creating community-specific word embeddings. All comments from a given subreddit were first concatenated into a single document, separated by 5 dummy tokens so adjacent comments did not influence the linguistic contexts of the first and last words. Following \citet{hamilton2016inducing}, word co-occurrence matrices for each subreddit were created with a symmetric context window of 4 words and reweighted using positive pointwise mutual information (PPMI) with context distribution smoothing $c = 0.75$ \cite{levy2015improving}. The dimensionality of each word embedding was then reduced to 100 using SVD.  

After obtaining subreddit-specific word embeddings, we introduce a small set of seed words with positive and negative polarity. We used the same seed words as ~\citet{hamilton2016inducing} did for Twitter, another social media platform (Table~\ref{tbl:sentiment_seed}). \textsc{SentProp} runs a series of random walks from both the positive and negative seed words, and the resulting sentiment value for each word is based on the probabilities that the word was hit by the positive random walk versus the negative one. We used \textsc{SentProp}'s default parameters for the Reddit lexicon induction portion of their paper, setting $\beta = 0.9$ and $K=25$, where $K$ is the number of nearest neighbors in the semantic space to which edges are drawn in graph construction. A higher $\beta$ favors similar labels for neighbors and a lower $\beta$ favors correct labels on seed words. We induced sentiment for the top 5000 words by frequency in each subreddit.

\begin{table*}
\centering
\footnotesize
\begin{tabular}
{|m{7cm}||m{7cm}|}
\hline
User-based Clusters & Text-based Clusters \\
\hline
\textbf{femalefashionadvice} \textbf{askwomen} \textbf{xxfitness} \textbf{trollxchromosomes} \textbf{actuallesbians} weddingplanning makeupaddiction justnomil skincareaddiction raisedbynarcissists dogs childfree vegan parenting running teachers unresolvedmysteries & \textbf{askwomen} \textbf{actuallesbians} \textbf{askmen} \textbf{askgaybros} \textbf{trollxchromosomes} suicidewatch justnomil deadbedrooms babybumps seduction raisedbynarcissists dogs childfree casualiama legaladvice parenting foreveralone dating\_advice teachers polyamory \\
\hline
\textbf{mensrights} sandersforpresident changemyview neutralpolitics forwardsfromgrandma the\_donald anarchism economics atheism subredditdrama & \textbf{mensrights} thathappened teenagers forwardsfromgrandma niceguys blackpeopletwitter roastme trashy facepalm photoshopbattles outoftheloop 4chan cringe\\
\hline
\textbf{askgaybros} \textbf{askmen} suicidewatch teenagers sex bodybuilding depression seduction offmychest ama advice foreveralone dating\_advice tinder polyamory & \textbf{xxfitness} fatlogic loseit keto cooking vegan running bodybuilding\\
\hline
\textbf{malefashionadvice} houston cooking churning entrepreneur seattle financialindependence investing travel homeimprovement jobs photography homebrewing bicycling personalfinance & \textbf{femalefashionadvice} \textbf{malefashionadvice} weddingplanning makeupaddiction skincareaddiction sneakers streetwear asianbeauty fashionreps  \\
\hline
\end{tabular}
\caption{The gendered subreddits and a subset of the subreddits that occur in the same clusters as them.}
\label{tbl:cluster}
\end{table*}

Adjusting the parameters $\beta$ and $K$ did not change our main conclusions or observations. Lowering $\beta$ from 0.9 to as far as 0.5 shrinks the overall range of sentiment values from -3 to 3 to about -2 to 2. Words with neutral sentiment tend to be slightly more positive or negative with lower values of $\beta$, but words with the highest polarities are consistent. Sentiment scores also remain steady when varying $K$. With $\beta=0.9$, the Pearson correlation of sentiment scores between $K=25$ and $K=15$ is 0.9183 ($p < 0.001$) and 0.9668 between $K=25$ and $K=35$ ($p < 0.001$) for r/xxfitness. 

We standardized values for each word to have zero mean and unit variance. The resulting sentiment vectors were then an array of negative and positive values corresponding to sentiment, averaged over 50 bootstrap-sampled runs. Each index in these vectors maps to a word in the vocabulary, which is the union of all subreddits' vocabularies. If a word's sentiment was not induced in a certain subreddit, its sentiment value is set to a neutral zero.

\subsection{Metrics}

We performed agglomerative clustering on all 400 subreddits' user- and text-based representations to see where gendered subreddits' users and content are situated within Reddit. We fixed the number of clusters to 20 and compared cluster sets provided by different representations by calculating their adjusted mutual information (AMI), where the possible range of values is 0 for random cluster and 1 for identical ones \cite{vinh2010information}. To further compare the different representations, we calculated the Spearman correlation between subreddits' pairwise similarities. We identify misalignments as subreddit pairs that have high similarity for one representation but low similarity for another.

We also implemented the misalignment identification method proposed by \citet{datta2017identifying}. This method subtracts two pairwise similarity rank matrices created by two different representation types, such as text and user, and z-score normalizes the difference matrix's columns and rows. Values in the final misalignment matrix are called $z^2$-scores. Pairs of subreddits with a high positive $z^2$-score have a higher similarity than expected with the first representation compared to the second, while a large negative $z^2$-score signifies the opposite. 

\section{Analysis}

\subsection{Text \& Users} \label{text_users}

The clusterings of user- and text-based representations are similar, with an AMI of 0.5610, and the text-based clusterings are more topically coherent. For example, r/femalefashionadvice and r/malefashionadvice are in the same text-based cluster but different user-based clusters, since both are about fashion but cater towards different genders. Subsets of the clusters in which gendered subreddits appear can be found in Table \ref{tbl:cluster}. The feminine subreddits are all in the same user-based cluster, while the masculine ones are more scattered. Female Reddit users may find themselves pushed into this cluster because of an overall predominant masculine culture throughout the platform which can be hostile to women \cite{massanari2017gamergate}. A majority of the gendered subreddits occur in the text-based cluster containing those related to personal topics, such as families and relationships. These clusters situate our gendered communities based on \textit{what} they talk about and \textit{who} is talking, and changing the representation type alters the perceived geography of Reddit.

\begin{figure}[!t]
    \centering
    \includegraphics[width=0.44\textwidth]{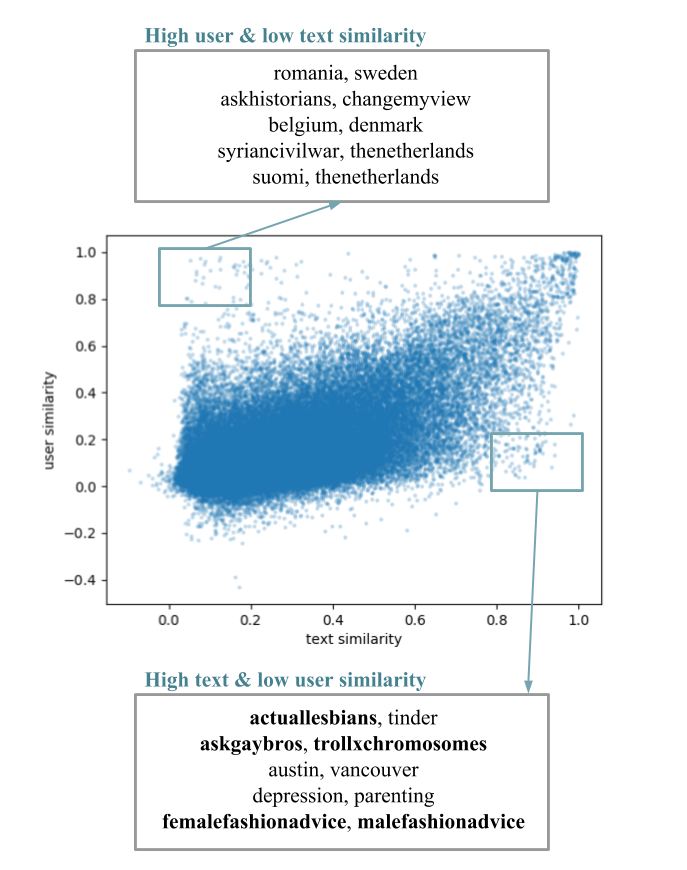}
    \caption{Subreddit similarities, with text similarities on the horizontal axis and user similarities on the vertical axis. Each point represents a pair of subreddits. The listed subreddit pairs are examples of outliers, where one type of similarity is very small ($< 0.2$) and another is large ($> 0.8$).}
    \label{fig:sim_corr}
\end{figure}

The Spearman correlation between text and user vectors' pairwise similarities is 0.549 ($p < 0.0001$), plotted in Figure~\ref{fig:sim_corr}. The vast majority of the pairs with high user and low text similarity are those pertaining to European countries, where comments are in different languages. The pairs with high text and low user similarity include subreddits pertaining to cities, as well as those divided by gender, sexual orientation, or personal topics. Thus, user demographics are important motivators for community formation on Reddit. User-based similarities for gendered subreddits tend to vary based on whether they cater towards the same gender, though some subreddits act as bridges between them: r/askmen and r/askwomen have a user similarity of 0.417, which is above the average among gendered subreddits (0.2880). The structure of these communities facilitates this by encouraging questions posted by users of any gender, and follow-up dialogue across groups accompanies the targeted gender's answers in the comments.

The misalignments identified by raw similarities are more intuitive than those identified by $z^2$-scores. Pairs in the top 20 with high text and low user similarity based on $z^2$-scores included understandable ones such as r/casualconversation-r/tinder, but also many pairs on very different topics such as r/bodybuilding-r/learningprogramming and r/makeupaddiction-r/legaladvice. Pairs in the top 20 with low text and high user similarity were easier to interpret and included r/truegaming-r/askmen and r/vegan-r/askscience. The $z^2$ score method normalizes any subreddit's skewed distribution of similarities. For example, country subreddits in general have lower text similarity to other subreddits since Reddit is mostly in English, and normalization causes inter-country similarities to match their expected similarity differences. Thus, normalization steps may reduce meaningful variation in how some subreddit similarities deviate from the norm of Reddit as a whole. 

\begin{table*}
\centering
\footnotesize
\begin{tabular}
{|l|c|l|c|}
\hline
Highest & Similarity & Lowest & Similarity\\ 
\hline
askmen, askwomen & 0.6702 & femalefashionadvice, mensrights & 0.1802 \\
askgaybros, askmen & 0.6144 & askwomen, malefashionadvice & 0.1876 \\
askwomen, trollxchromosomes & 0.6003 & malefashionadvice, trollxchromosomes & 0.2162\\
actuallesbians, trollxchromosomes & 0.5462 & malefashionadvice, mensrights & 0.2170 \\
askgaybros, askwomen & 0.5310 & mensrights, xxfitness & 0.2181 \\
\hline
\end{tabular}
\caption{Highest and lowest sentiment similarities between gendered subreddits, with a mean of 0.3701.}
\label{tbl:high_low_sent}
\end{table*}
 
\begin{table}
\centering
\begin{tabular}
{|l|l|l|}
\hline
& Spearman $\rho$ & $p$\\
\hline
text, user & 0.4268 & $< 0.01$ \\
text, sentiment & 0.6371 & $< 0.0001$  \\
user, sentiment & 0.4219 & $= 0.01$\\
\hline
\end{tabular}
\caption{Correlations of text, user, and sentiment representations using pairwise cosine similarity between nine gendered subreddits. Though sentiment and text are related, they provide different information.}
\label{tbl:correlations}
\end{table}

\subsection{Sentiment}

\begin{table*}
\centering
\footnotesize
\resizebox{0.9\linewidth}{!}{
\begin{tabular}
{|cc|cc|cc|cc|}
\hline
\multicolumn{2}{|c|}{TrollXChromosomes} & \multicolumn{2}{c|}{FemaleFashionAdvice} & \multicolumn{2}{c|}{MaleFashionAdvice} & \multicolumn{2}{c|}{MensRights} \\
\hline
Positive & Negative & Positive & Negative  & Positive & Negative & Positive & Negative \\
\hline
lovely & infections 	& gorgeous & gross 	& sweet & gross 	& wonderful & vile \\
gorgeous & yeast 	& adore & marks 	& beautiful & annoying & favorite & evil \\
beautiful & pain 	& lovely & blood 	& cool & dirty & excellent & disgusting \\
wonderful & dealing 	& stunning & rough   & those & stupid      & watch & horribly \\
congratulations & mess 	& thursdays & painful 	& vouch & armpits & enjoyed & cruel \\
fabulous & horrific 	& mondays & worse 	& plus & shitty & honey & hating     \\
congrats & painful 	& tuesdays & messed  	& dig & crap & clip & misogynistic \\
yay & minor 	& killer & causing 	& perfect & garbage & enjoying & hateful \\
d & uti 	& loooove & horribly 	& makes & crappy & fan & sexist \\
lt3 & infection 	& fabulous & poor 	& interesting & sweaty & episode & bigots \\
\hline
\end{tabular}}
\caption{Most positive and negative non-seed words for a selected set of subreddits.}
\label{tbl:gender_sent}
\end{table*}

\begin{table*}
\centering
\small
\begin{tabular}
{|c|c|c|c|}
\hline
Word & Variance & Positive Subreddits & Negative Subreddits \\
\hline
sounds & 2.775 & femalefashionadvice, actuallesbians & askgaybros, askmen \\
smells & 2.652 & actuallesbians, malefashionadvice & askgaybros, askmen \\
hilarious & 2.547 & askwomen, actuallesbians & askgaybros, malefashionadvice \\
absolutely & 2.231 & femalefashionadvice, malefashionadvice & askwomen, askmen \\
obsessed & 2.094 & askwomen, femalefashionadvice & mensrights, askmen \\
sharp & 2.087 & actuallesbians, femalefashionadvice & xxfitness, trollxchromosomes \\
\hline
\end{tabular}
\caption{Words with greatest variance in sentiment across all gendered subreddits, along with the subreddits in which they are the most positive and most negative.}
\label{tbl:topvariance}
\end{table*}

We calculated the pairwise cosine similarities between each of the nine explicitly gendered subreddits using text, user, and sentiment community representations. Table \ref{tbl:correlations} shows the resulting Spearman correlation between these representations. Sentiment representations correlate more strongly with those of text, which could be explained by how they are both linguistically motivated. However, this correlation is far from 1, suggesting that sentiment representations capture some aspects of communities that weighted word counts do not. 

The pairs of subreddits with highest and lowest sentiment similarity can be found in Table \ref{tbl:high_low_sent}. Some of the highest similarities are between subreddits oriented towards the same gender, and while some of the lowest are between those of different genders, but there are several exceptions. Therefore, sentiment does not divide itself evenly based on gender. The high sentiment similarity between r/askmen and r/askwomen misaligns with their low text similarity (0.2874) and near average user similarity (0.4168).  Another outlier across text, user, and sentiment similarities is r/actuallesbians and r/trollxchromosomes, which have high user similarity (0.8856), above average sentiment similarity (0.5462), and high text similarity (0.9415). 
 
The most positive and negative non-seed words in each subreddit are consistent with the concepts we expect to be relevant to them (some examples\footnote{The token \textit{lt3} is a punctuation-stripped HTML heart. Similarly, \textit{d} is often a happy emoji \textit{:D}.} in Table \ref{tbl:gender_sent}.). Many negative words in r/trollxchromosomes and r/femalefashionadvice revolve around pain and health, while those in r/mensrights refer to gender bias (within the top fifteen most negative words are \textit{misandrist} and \textit{manhating}). The most positive words in r/xxfitness are similar to those in other subreddits since many are adjectives such as \textit{great} and \textit{fun}, but its most negative words almost entirely focus on physical ailments, such as \textit{flu}, \textit{infection} and \textit{headache}. The words \textit{brothers} and \textit{brother} have highest sentiment in r/mensrights compared to other gendered subreddits, suggesting that this community values masculine solidarity. Likewise, even though the words \textit{troll} and \textit{trolls} are predominantly negative, they have high positive sentiment in r/trollxchromosomes, as users in this community have re-appropriated these terms to refer positively to themselves.\footnote{The full lexicons can be found in our Github repo \href{https://github.com/lucy3/reddit-sent/tree/master/logs/socialsent_lexicons_ppmi_svd_top5000}{here}.} 

Subreddits with high text similarity such as r/malefashionadvice and r/femalefashionadvice still contain distinct cultures. The highly positive words in r/femalefashionadvice reflect their custom of referencing daily outfits using days of the week, while users on r/malefashionadvice do not follow this format. The expressive elongation in r/femalefashionadvice's highly positive \textit{loooove} has previously been shown to be a female marker \cite{bamman2014gender,rao2010classifying}. Sentiment is a helpful but sometimes superficial metric for determining community values, and its interpretation is best understood in context with topic and users. For example, \textit{men} is most negative in r/mensrights compared to other gendered subreddits, but that does not mean these users dislike men, since the opposite is actually the case. The strong negativity here is instead associated with how their discussions center on injustices towards men. 

Table \ref{tbl:topvariance} shows the words with the highest variance in sentiment with the subreddits in which they have the most positive and negative polarity. Much of the cross-community variation in sentiment is likely due to polysemy, where different senses of a given word are predominant in different communities. For example, when calculating the sentiment of \textit{sick} (13th highest variance) in each subreddit's semantic space, \textsc{SentProp} encounters neighbors such as \textit{nauseous} (r/xxfitness), \textit{disgusting} (r/mensrights), or \textit{dope} (r/malefashionadvice). 
\begin{itemize}
\setlength\itemsep{0em}
{\small
\item[] \textit{i never thought id see sitting on a tricycle look so badass looks \textbf{sick} love it\footnote{Examples transcribed as they appear after preprocessing.}} (r/malefashionadvice)
\item[] ...\textit{what kind of \textbf{sick} twisted person could do that to another human being truly disgusting what some people are willing to do to others} (r/mensrights)
}
\end{itemize}


Furthermore, words may adopt the sentiment polarities that reflect the overall discourse style. In r/femalefashionadvice, seemingly negative words such as \textit{jealous} (10th highest variance) take on a positive meaning as they are used to compliment the original poster:
\begin{itemize}
\setlength\itemsep{0em}
{\small
\item[] \textit{that is amazing congratulations i am so \textbf{jealous}} (r/femalefashionadvice)}
\end{itemize}

This relationship with overall discourse style may be even more pronounced for judgment-related words such as \textit{sounds}, whose polarity reflects whether communities tend to use it to evaluate other users or entities positively or negatively. 
\begin{itemize}
\setlength\itemsep{0em}
{\small
\item[] ...\textit{that \textbf{sounds} like a really polite and productive way to deal with the gift issues} (r/femalefashionadvice)
\item[] \textit{\textbf{sounds} like you just have shitty friends} (r/askgaybros)}
\end{itemize}

\begin{figure}[t]
    \centering
    \includegraphics[scale=0.49]{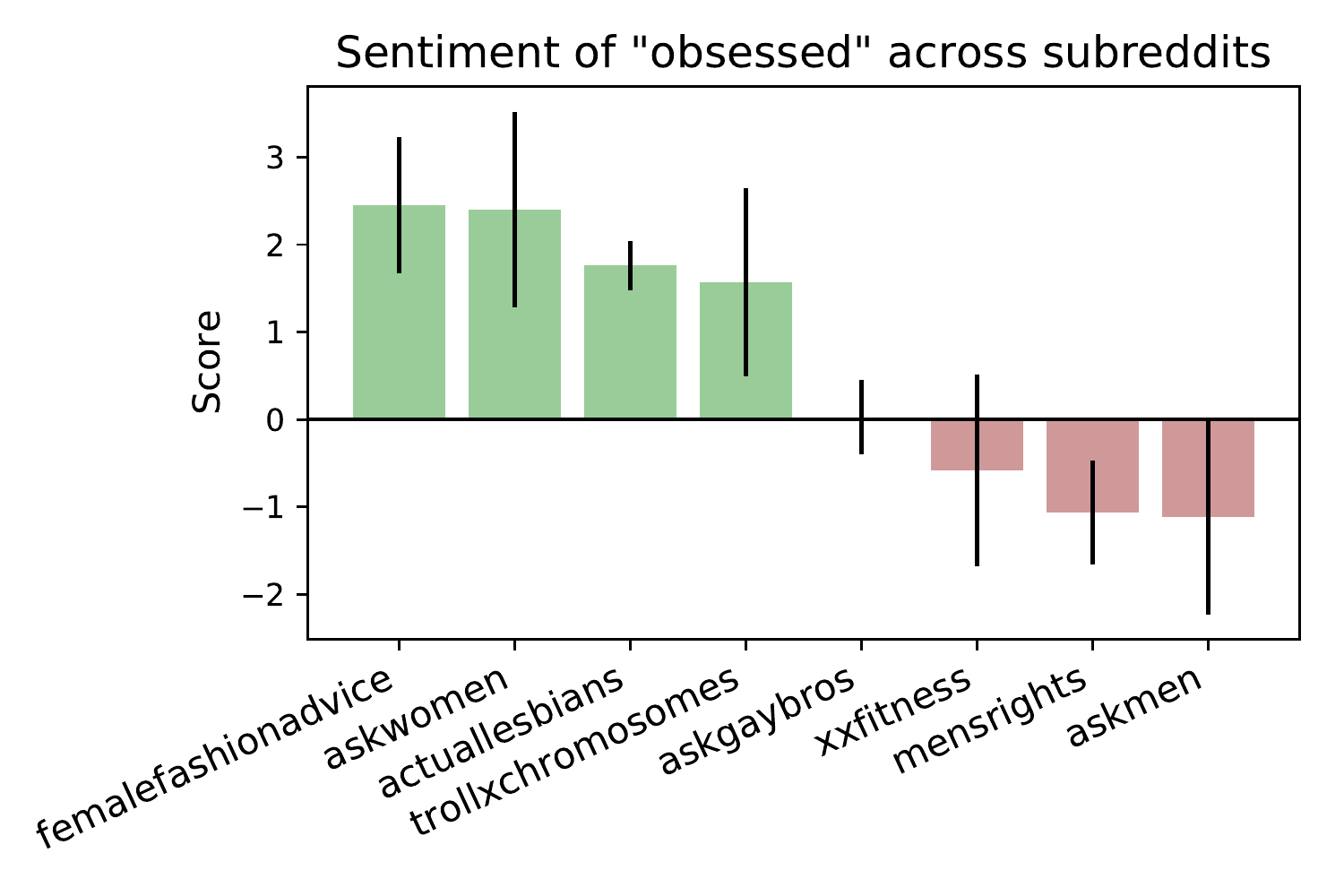}
    \caption{Words with high variance across communities such as \textit{obsessed} can be strongly positive or negative.}%
    \label{fig:sent_obsessed}%
\end{figure}

Finally, many affective differences emerge depending on whether a subreddit's users talk about their own feelings, beliefs, and passions (personal) or make claims about other people's mental states (impersonal). In particular, sentiment of words such as \textit{jealous} or \textit{obsessed} varies depending on whether one uses it to describe themselves or somebody else (Figure \ref{fig:sent_obsessed}).
\begin{itemize}
\setlength\itemsep{0em}
{\small
\item[] ...\textit{recently became \textbf{obsessed} with this podcast this is super cool } (r/trollxchromosomes)
\item[] \textit{so sad that so many female teachers are feminist \textbf{obsessed}...} (r/mensrights)}
\end{itemize}

\citet{volkova2013exploring} studied sentiment variation using Twitter data, but treated gender as a binary variable of male versus female. They listed examples with large gender differences: \textit{weakness} is used positively by women and negatively by men, while \textit{overdressed} is used positively by men and negatively by women. Their most polarized words, with hashtags removed, do not split according to a binary in our subreddits. We observe diverse language styles: \textit{weakness} is strongly negative in r/xxfitness (-2.0634 $\pm$ 0.5577) but positive in r/actuallesbians (0.9112 $\pm$ 0.2100), and the sentiment score of \textit{overdressed} is similar in r/malefashionadvice (-0.6016 $\pm$ 0.5273) and r/femalefashionadvice (-0.6696 $\pm$ 0.8512). This implies that though gender can be a helpful variable for improving sentiment analysis, its expression is not fixed across multiple contexts. 

\begin{figure}[t]
    \centering
    \includegraphics[scale=0.49]{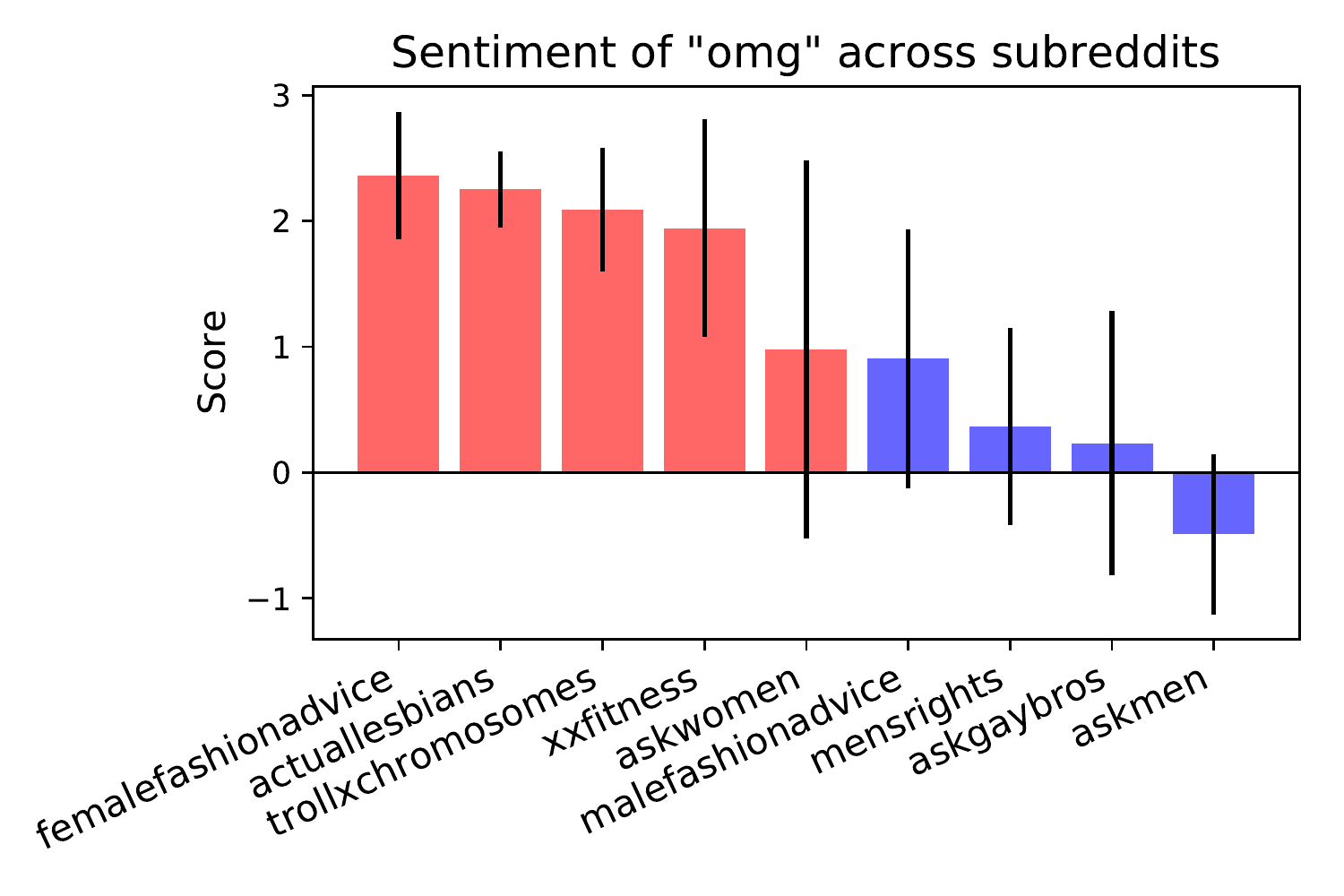}
    \caption{Average sentiment scores and standard deviations of ``omg" in explicitly-gendered subreddits, sorted in decreasing order, based on 50 \textsc{SentProp} runs. Women-oriented subreddits are marked in red, and men-oriented are in blue. }%
    \label{fig:sent_omg}%
\end{figure}

The word \textit{omg} seems to be used in far broader contexts and have a substantially different meaning than its origin phrase, \textit{oh my god}. In Figure \ref{fig:sent_omg}, the sentiment of \textit{omg} is highest among the five women-oriented subreddits, and lower in the men-oriented subreddits. This finding is consistent with prior results that show that forms predominant in computer-mediated communication are more commonly used by and highly associated with women \cite{bamman2014gender,carpenter2017real}. However, these results have implications beyond just associating forms such as \textit{omg} with women. In particular, \textit{omg} is not a filler word devoid of meaning in women-oriented communities. Rather, it conveys highly positive affect and may also indicate cooperativeness and engagement in a conversation. It seems to not play the same role in men-oriented communities, where \textit{omg} is used frequently in indirect quotes of others' speech. 
\begin{itemize}
	\setlength\itemsep{0em}
	{\small
		\item[] \textit{\textbf{omg} love her did you see that shoeprinted dress she had on this weeks episode  } (r/femalefashionadvice)
		\item[] \textit{...if anything it was usually \textbf{omg} you two would have the most beautiful babies...} (r/askmen)}
\end{itemize}

\begin{figure}[t]
    \centering
    \includegraphics[scale=0.49]{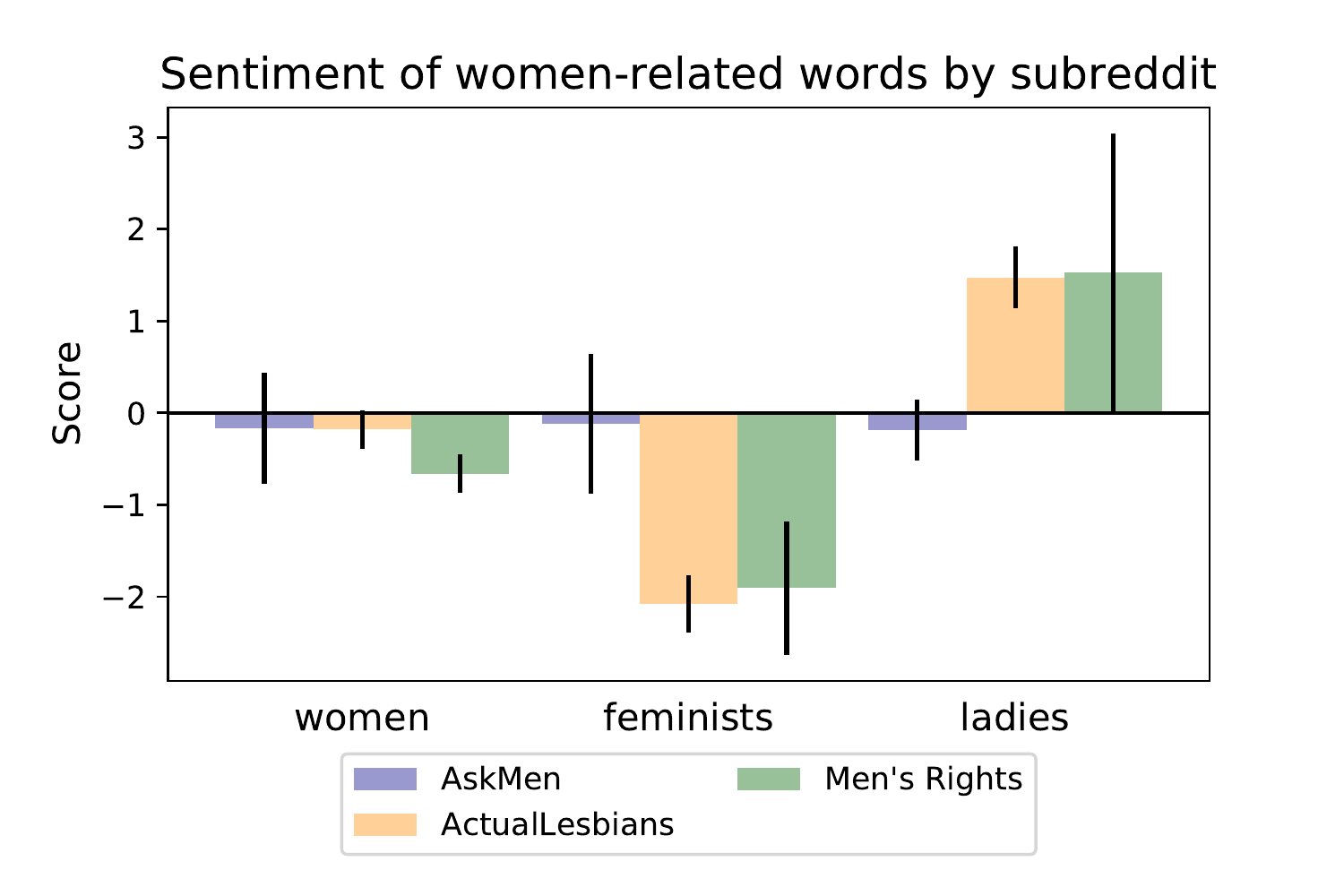}
    \caption{Sentiment scores of the words ``women", ``feminists", and ``ladies" across three subreddits, with error bars showing the standard deviation of 50 bootstrap-sampled \textsc{SentProp} runs.}%
    \label{fig:sent_women}%
\end{figure}

Sentiment-based representations can also detect words that may be denotationally similar but have different social meanings due to repeated associations with certain beliefs and stereotypes. Figure \ref{fig:sent_women} demonstrates community variation in the affective meaning of the denotationally similar terms \textit{women} and \textit{ladies} and the related but semantically distinct \textit{feminists} across three subreddits. This variation demonstrates the potential for denotationally similar terms to acquire community-dependent connotational and affective meanings. 

Unsurprisingly, \textit{feminists} is highly negative in r/mensrights. However, it is similarly negative in r/actuallesbians, while neutral in r/askmen. Members of r/actuallesbians tend to not disapprove of feminists in general; instead, much of the discourse that includes this word focuses on the perceived exclusivity of many feminist movements towards LGBTQ individuals. 
\begin{itemize}
\setlength\itemsep{0em}
{\small
\item[] \textit{...didnt realize that transphobia was such an organized and politically influential problem especially from some \textbf{feminists} and not just old white guys...} (r/actuallesbians)}
\end{itemize}

The words \textit{women} and \textit{ladies} are associated with many distinct social meanings. For example, \textit{women} is seen as both a neutral and cold label, while \textit{ladies} can be seen as traditional, patronizing, and sexual \cite{cralley2005lady,friedman2013}. More recently, \textit{ladies} has also been reclaimed as an age-agnostic label popular with modern feminists \cite{friedman2013}.

Both \textit{women} and \textit{ladies} are used in r/actuallesbians to name the target group of romantic and sexual attraction. However, the much more positive sentiment of \textit{ladies} may be due to its additional meaning as an in-group label.
\begin{itemize}
	\setlength\itemsep{0em}
	{\small
		\item[] \textit{its a pretty slow reddit but \textbf{ladies} do say hello and interact} (r/actuallesbians)
        }
\end{itemize}

Even though \textit{ladies} also has positive sentiment in r/mensrights, the word is used very differently there. Instead of being used to sexualize women, \textit{ladies} was far more commonly written in a patronizing manner. 

\begin{itemize}
	\setlength\itemsep{0em}
	{\small
		\item[] \textit{...\textbf{ladies} this is what equality looks like time to give up some of your numerous privileges} (r/mensrights)
		}
\end{itemize}

This reveals one limitation of a sentiment-only approach to analyzing sociolinguistic variation. The patronizing and sometimes sarcastic usage of \textit{ladies} is a complex phenomenon that cannot be easily captured by methods based on vector space models. \textsc{SentProp} mistakenly arrives at a positive polarity for \textit{ladies} in r/mensrights, although its high standard deviation hints at some underlying source of inconsistency. 

\section{Conclusion} 

Sentiment representations are useful for understanding variation both on a broad scale as well as among specific lexemes, particularly when combined with in-depth qualitative analyses. We focused on only explicitly gendered subreddits, but other subreddits can also be implicitly gendered. From the user-based clusters, we may be able to infer that subreddits like r/weddingplanning and r/makeupaddiction have mostly feminine users. In the future, we would like to situate our sentiment analysis of gendered subreddits in the larger context of Reddit. For example, comparing r/xxfitness with its fitness-related neighbors may allow a understanding of how explicitly targeting some demographic changes words' sentiment. 

Sentiment is a salient semantic dimension, but it may also be illuminating to define subreddits along some other dimension, such as arousal, concreteness, or various emotions \cite{brysbaert2014concreteness, warriner2013norms, mohammad2010emotions}. Rarely are subreddit communities redundant. Though two subreddits may align with high similarities with some type of representation, they should differ in some other one. Still, a comparison of sentiment and emotion could result in strong alignment; negativity underlies anger and sadness, while positivity is fundamental to happiness and surprise. 

\section{Acknowledgements} 

We would like to thank Chris Potts, Will Hamilton, and Bill MacCartney for their helpful ideas and comments. Additional thanks to Will for Reddit data. 

\bibliography{naaclhlt2018}

\begin{thebibliography}{42}
\expandafter\ifx\csname natexlab\endcsname\relax\def\natexlab#1{#1}\fi

\bibitem[{Althoff et~al.(2014)Althoff, Danescu-Niculescu-Mizil, and
  Jurafsky}]{althoff2014ask}
Tim Althoff, Cristian Danescu-Niculescu-Mizil, and Dan Jurafsky. 2014.
\newblock How to ask for a favor: A case study on the success of altruistic
  requests.
\newblock In \emph{Proceedings of International Conference on Web and Social
  Media}.

\bibitem[{Argamon et~al.(2007)Argamon, Koppel, Pennebaker, and
  Schler}]{argamon2007mining}
Shlomo Argamon, Moshe Koppel, James~W Pennebaker, and Jonathan Schler. 2007.
\newblock Mining the blogosphere: Age, gender and the varieties of
  self-expression.
\newblock \emph{First Monday}, 12(9).

\bibitem[{Bamman et~al.(2014)Bamman, Eisenstein, and
  Schnoebelen}]{bamman2014gender}
David Bamman, Jacob Eisenstein, and Tyler Schnoebelen. 2014.
\newblock Gender identity and lexical variation in social media.
\newblock \emph{Journal of Sociolinguistics}, 18(2):135--160.

\bibitem[{Brysbaert et~al.(2014)Brysbaert, Warriner, and
  Kuperman}]{brysbaert2014concreteness}
Marc Brysbaert, Amy~Beth Warriner, and Victor Kuperman. 2014.
\newblock Concreteness ratings for 40 thousand generally known english word
  lemmas.
\newblock \emph{Behavior research methods}, 46(3):904--911.

\bibitem[{Burger et~al.(2011)Burger, Henderson, Kim, and
  Zarrella}]{burger2011discriminating}
John~D Burger, John Henderson, George Kim, and Guido Zarrella. 2011.
\newblock Discriminating gender on twitter.
\newblock In \emph{Proceedings of the conference on empirical methods in
  natural language processing}, pages 1301--1309. Association for Computational
  Linguistics.

\bibitem[{Butler(1988)}]{butler1988performative}
Judith Butler. 1988.
\newblock Performative acts and gender constitution: An essay in phenomenology
  and feminist theory.
\newblock \emph{Theatre journal}, 40(4):519--531.

\bibitem[{Carpenter et~al.(2017)Carpenter, Preotiuc-Pietro, Flekova, Giorgi,
  Hagan, Kern, Buffone, Ungar, and Seligman}]{carpenter2017real}
Jordan Carpenter, Daniel Preotiuc-Pietro, Lucie Flekova, Salvatore Giorgi,
  Courtney Hagan, Margaret~L Kern, Anneke~EK Buffone, Lyle Ungar, and Martin~EP
  Seligman. 2017.
\newblock Real men don’t say “cute” using automatic language analysis to
  isolate inaccurate aspects of stereotypes.
\newblock \emph{Social Psychological and Personality Science}, 8(3):310--322.

\bibitem[{Cralley and Ruscher(2005)}]{cralley2005lady}
Elizabeth~L Cralley and Janet~B Ruscher. 2005.
\newblock Lady, girl, female, or woman: Sexism and cognitive busyness predict
  use of gender-biased nouns.
\newblock \emph{Journal of Language and Social Psychology}, 24(3):300--314.

\bibitem[{Danescu-Niculescu-Mizil et~al.(2013)Danescu-Niculescu-Mizil, West,
  Jurafsky, Leskovec, and Potts}]{danescu2013no}
Cristian Danescu-Niculescu-Mizil, Robert West, Dan Jurafsky, Jure Leskovec, and
  Christopher Potts. 2013.
\newblock No country for old members: User lifecycle and linguistic change in
  online communities.
\newblock In \emph{Proceedings of the 22nd international conference on World
  Wide Web}, pages 307--318. ACM.

\bibitem[{Datta et~al.(2017)Datta, Phelan, and Adar}]{datta2017identifying}
Srayan Datta, Chanda Phelan, and Eytan Adar. 2017.
\newblock \href {https://doi.org/10.1145/3134672} {Identifying misaligned
  inter-group links and communities}.
\newblock \emph{Proc. ACM Hum.-Comput. Interact.}, 1(CSCW):37:1--37:23.

\bibitem[{Eckert(1989)}]{eckert1989jocks}
Penelope Eckert. 1989.
\newblock \emph{Jocks and burnouts: Social categories and identity in the high
  school}.
\newblock Teachers College Press.

\bibitem[{Eckert(2006)}]{eckert2006communities}
Penelope Eckert. 2006.
\newblock Communities of practice.
\newblock \emph{Encyclopedia of language and linguistics}, 2(2006):683--685.

\bibitem[{Eckert(2012)}]{eckert2012three}
Penelope Eckert. 2012.
\newblock Three waves of variation study: The emergence of meaning in the study
  of sociolinguistic variation.
\newblock \emph{Annual review of Anthropology}, 41:87--100.

\bibitem[{Eckert and McConnell-Ginet(1992)}]{eckert1992think}
Penelope Eckert and Sally McConnell-Ginet. 1992.
\newblock Think practically and look locally: Language and gender as
  community-based practice.
\newblock \emph{Annual review of anthropology}, 21(1):461--488.

\bibitem[{Friedman(2013)}]{friedman2013}
Ann Friedman. 2013.
\newblock \href
  {https://newrepublic.com/article/112188/how-word-lady-has-evolved} {Hey
  "ladies": The unlikely revival of a fusty old label}.
\newblock \emph{The New Republic}.

\bibitem[{Halko et~al.(2011)Halko, Martinsson, and Tropp}]{halko2011finding}
Nathan Halko, Per-Gunnar Martinsson, and Joel~A Tropp. 2011.
\newblock Finding structure with randomness: Probabilistic algorithms for
  constructing approximate matrix decompositions.
\newblock \emph{SIAM review}, 53(2):217--288.

\bibitem[{Hall(2009)}]{hall2009boys}
Kira Hall. 2009.
\newblock Boys’ talk: Hindi, moustaches and masculinity in new delhi.
\newblock In \emph{Gender and spoken interaction}, pages 139--162. Springer.

\bibitem[{Hamilton et~al.(2016)Hamilton, Clark, Leskovec, and
  Jurafsky}]{hamilton2016inducing}
William~L Hamilton, Kevin Clark, Jure Leskovec, and Dan Jurafsky. 2016.
\newblock Inducing domain-specific sentiment lexicons from unlabeled corpora.
\newblock In \emph{Proceedings of the 2016 Conference on Empirical Methods in
  Natural Language Processing}, pages 595--605.

\bibitem[{Hamilton et~al.(2017)Hamilton, Zhang, Danescu-Niculescu-Mizil,
  Jurafsky, and Leskovec}]{hamilton2017loyalty}
William~L Hamilton, Justine Zhang, Cristian Danescu-Niculescu-Mizil, Dan
  Jurafsky, and Jure Leskovec. 2017.
\newblock Loyalty in online communities.
\newblock In \emph{Proceedings of the International AAAI Conference on Weblogs
  and Social Media}, volume 2017, page 540. NIH Public Access.

\bibitem[{Herring and Paolillo(2006)}]{herring2006gender}
Susan~C Herring and John~C Paolillo. 2006.
\newblock Gender and genre variation in weblogs.
\newblock \emph{Journal of Sociolinguistics}, 10(4):439--459.

\bibitem[{Jaech et~al.(2015)Jaech, Zayats, Fang, Ostendorf, and
  Hajishirzi}]{jaech2015talking}
Aaron Jaech, Victoria Zayats, Hao Fang, Mari Ostendorf, and Hannaneh
  Hajishirzi. 2015.
\newblock Talking to the crowd: What do people react to in online discussions?
\newblock In \emph{Proceedings of the 2015 Conference on Empirical Methods in
  Natural Language Processing}, pages 2026--2031.

\bibitem[{Kumar et~al.(2018)Kumar, Hamilton, Leskovec, and
  Jurafsky}]{kumar2018community}
Srijan Kumar, William~L Hamilton, Jure Leskovec, and Dan Jurafsky. 2018.
\newblock Community interaction and conflict on the web.
\newblock In \emph{Proceedings of the 2018 World Wide Web Conference on World
  Wide Web}, pages 933--943. International World Wide Web Conferences Steering
  Committee.

\bibitem[{Levy et~al.(2015)Levy, Goldberg, and Dagan}]{levy2015improving}
Omer Levy, Yoav Goldberg, and Ido Dagan. 2015.
\newblock Improving distributional similarity with lessons learned from word
  embeddings.
\newblock \emph{Transactions of the Association for Computational Linguistics},
  3:211--225.

\bibitem[{Martin(2017)}]{martin2017community2vec}
Trevor Martin. 2017.
\newblock community2vec: Vector representations of online communities encode
  semantic relationships.
\newblock In \emph{Proceedings of the Second Workshop on NLP and Computational
  Social Science}, pages 27--31.

\bibitem[{Massanari(2017)}]{massanari2017gamergate}
Adrienne Massanari. 2017.
\newblock \# gamergate and the fappening: How reddit’s algorithm, governance,
  and culture support toxic technocultures.
\newblock \emph{New Media \& Society}, 19(3):329--346.

\bibitem[{Mendoza-Denton(1996)}]{mendoza1996muy}
Norma Mendoza-Denton. 1996.
\newblock ‘muy macha’: Gender and ideology in gang-girls’ discourse about
  makeup.
\newblock \emph{Ethnos}, 61(1-2):47--63.

\bibitem[{Mohammad and Turney(2010)}]{mohammad2010emotions}
Saif~M Mohammad and Peter~D Turney. 2010.
\newblock Emotions evoked by common words and phrases: Using mechanical turk to
  create an emotion lexicon.
\newblock In \emph{Proceedings of the NAACL HLT 2010 workshop on computational
  approaches to analysis and generation of emotion in text}, pages 26--34.
  Association for Computational Linguistics.

\bibitem[{Mulac et~al.(2001)Mulac, Bradac, and Gibbons}]{mulac2001empirical}
Anthony Mulac, James~J Bradac, and Pamela Gibbons. 2001.
\newblock Empirical support for the gender-as-culture hypothesis: An
  intercultural analysis of male/female language differences.
\newblock \emph{Human Communication Research}, 27(1):121--152.

\bibitem[{Newell et~al.(2016)Newell, Jurgens, Saleem, Vala, Sassine, Armstrong,
  and Ruths}]{newell2016user}
Edward Newell, David Jurgens, Haji~Mohammad Saleem, Hardik Vala, Jad Sassine,
  Caitrin Armstrong, and Derek Ruths. 2016.
\newblock User migration in online social networks: A case study on reddit
  during a period of community unrest.
\newblock In \emph{Tenth International AAAI Conference on Web and Social
  Media}.

\bibitem[{Newman et~al.(2008)Newman, Groom, Handelman, and
  Pennebaker}]{newman2008gender}
Matthew~L Newman, Carla~J Groom, Lori~D Handelman, and James~W Pennebaker.
  2008.
\newblock Gender differences in language use: An analysis of 14,000 text
  samples.
\newblock \emph{Discourse Processes}, 45(3):211--236.

\bibitem[{Nguyen et~al.(2014)Nguyen, Trieschnigg, Do{\u{g}}ru{\"o}z, Gravel,
  Theune, Meder, and De~Jong}]{nguyen2014gender}
Dong Nguyen, Dolf Trieschnigg, A~Seza Do{\u{g}}ru{\"o}z, Rilana Gravel,
  Mari{\"e}t Theune, Theo Meder, and Franciska De~Jong. 2014.
\newblock Why gender and age prediction from tweets is hard: Lessons from a
  crowdsourcing experiment.
\newblock In \emph{Proceedings of COLING 2014, the 25th International
  Conference on Computational Linguistics: Technical Papers}, pages 1950--1961.

\bibitem[{Pavalanathan et~al.(2017)Pavalanathan, Fitzpatrick, Kiesling, and
  Eisenstein}]{pavalanathan2017multidimensional}
Umashanthi Pavalanathan, Jim Fitzpatrick, Scott Kiesling, and Jacob Eisenstein.
  2017.
\newblock A multidimensional lexicon for interpersonal stancetaking.
\newblock In \emph{Proceedings of the 55th Annual Meeting of the Association
  for Computational Linguistics (Volume 1: Long Papers)}, volume~1, pages
  884--895.

\bibitem[{Pedregosa et~al.(2011)Pedregosa, Varoquaux, Gramfort, Michel,
  Thirion, Grisel, Blondel, Prettenhofer, Weiss, Dubourg, Vanderplas, Passos,
  Cournapeau, Brucher, Perrot, and Duchesnay}]{scikit-learn}
F.~Pedregosa, G.~Varoquaux, A.~Gramfort, V.~Michel, B.~Thirion, O.~Grisel,
  M.~Blondel, P.~Prettenhofer, R.~Weiss, V.~Dubourg, J.~Vanderplas, A.~Passos,
  D.~Cournapeau, M.~Brucher, M.~Perrot, and E.~Duchesnay. 2011.
\newblock Scikit-learn: Machine learning in {P}ython.
\newblock \emph{Journal of Machine Learning Research}, 12:2825--2830.

\bibitem[{Rao et~al.(2010)Rao, Yarowsky, Shreevats, and
  Gupta}]{rao2010classifying}
Delip Rao, David Yarowsky, Abhishek Shreevats, and Manaswi Gupta. 2010.
\newblock Classifying latent user attributes in twitter.
\newblock In \emph{Proceedings of the 2nd international workshop on Search and
  mining user-generated contents}, pages 37--44. ACM.

\bibitem[{Rothe et~al.(2016)Rothe, Ebert, and
  Sch{\"u}tze}]{rothe2016ultradense}
Sascha Rothe, Sebastian Ebert, and Hinrich Sch{\"u}tze. 2016.
\newblock Ultradense word embeddings by orthogonal transformation.
\newblock In \emph{Proceedings of the 2016 Conference of the North American
  Chapter of the Association for Computational Linguistics: Human Language
  Technologies}, pages 767--777.

\bibitem[{Schler(2006)}]{schler2006effects}
Jonathan Schler. 2006.
\newblock Effects of age and gender on blogging.
\newblock In \emph{Proceedings of AAAI Symposium on Computational Approaches
  for Analyzing Weblogs, 2006}, pages 199--205.

\bibitem[{Vinh et~al.(2010)Vinh, Epps, and Bailey}]{vinh2010information}
Nguyen~Xuan Vinh, Julien Epps, and James Bailey. 2010.
\newblock Information theoretic measures for clusterings comparison: Variants,
  properties, normalization and correction for chance.
\newblock \emph{Journal of Machine Learning Research}, 11(Oct):2837--2854.

\bibitem[{Volkova et~al.(2013)Volkova, Wilson, and
  Yarowsky}]{volkova2013exploring}
Svitlana Volkova, Theresa Wilson, and David Yarowsky. 2013.
\newblock Exploring demographic language variations to improve multilingual
  sentiment analysis in social media.
\newblock In \emph{Proceedings of the 2013 Conference on Empirical Methods in
  Natural Language Processing}, pages 1815--1827.

\bibitem[{Warriner et~al.(2013)Warriner, Kuperman, and
  Brysbaert}]{warriner2013norms}
Amy~Beth Warriner, Victor Kuperman, and Marc Brysbaert. 2013.
\newblock Norms of valence, arousal, and dominance for 13,915 english lemmas.
\newblock \emph{Behavior research methods}, 45(4):1191--1207.

\bibitem[{Wong(2005)}]{wong2005reappropriation}
Andrew~D Wong. 2005.
\newblock The reappropriation of tongzhi.
\newblock \emph{Language in society}, 34(5):763--793.

\bibitem[{Yang and Eisenstein(2015)}]{yang2015putting}
Yi~Yang and Jacob Eisenstein. 2015.
\newblock Putting things in context: Community-specific embedding projections
  for sentiment analysis.
\newblock \emph{Arxiv-Social Media Intelligence}.

\bibitem[{Zhang et~al.(2017)Zhang, Hamilton, Danescu-Niculescu-Mizil, Jurafsky,
  and Leskovec}]{zhang2017community}
Justine Zhang, William~L Hamilton, Cristian Danescu-Niculescu-Mizil, Dan
  Jurafsky, and Jure Leskovec. 2017.
\newblock Community identity and user engagement in a multi-community
  landscape.
\newblock In \emph{Proceedings of the... International AAAI Conference on
  Weblogs and Social Media. International AAAI Conference on Weblogs and Social
  Media}, volume 2017, page 377. NIH Public Access.

\end{thebibliography}
\bibliographystyle{acl_natbib}

\end{document}